\documentclass[preprint,12pt, authoryear]{elsarticle}

\usepackage{amssymb}
\usepackage{times}  
\usepackage{helvet}  
\usepackage{courier}  
\usepackage{url}  
\usepackage{graphicx} 
\usepackage{natbib}  
\usepackage{caption} 
\usepackage{graphicx} 
\usepackage{algorithm}
\usepackage{algorithmic}
\usepackage{changes}
\usepackage{tabularx}
\usepackage{xcolor}         
\usepackage{amsmath}
\usepackage{caption}
\usepackage{subcaption}
\usepackage{amssymb}
\usepackage{wrapfig} 
\usepackage{multirow}
\usepackage{changes}
\usepackage{balance}
\newtheorem{definition}{Definition}
\newtheorem{remark}{Remark}
\newtheorem{proposition}{Proposition}

\journal{Journal}

\begin{document}

\begin{frontmatter}


\title{Human-Inspired Framework to Accelerate Reinforcement Learning}


\author[l1]{Ali Beikmohammadi \corref{cor1}}
\ead{beikmohammadi@dsv.su.se}
\author[l1]{Sindri Magnússon}
\ead{sindri.magnusson@dsv.su.se}
\affiliation[l1]{organization={Department of Computer and Systems Sciences},
            addressline={Stockholm University}, 
            postcode={SE-164 55}, 
            state={Stockholm},
            country={Sweden}}
\cortext[cor1]{Corresponding Author}

\begin{abstract}
Reinforcement learning (RL) is crucial for data science decision-making but suffers from sample inefficiency, particularly in real-world scenarios with costly physical interactions. This paper introduces a novel human-inspired framework to enhance RL algorithm sample efficiency. It achieves this by initially exposing the learning agent to simpler tasks that progressively increase in complexity, ultimately leading to the main task. This method requires no pre-training and involves learning simpler tasks for just one iteration. The resulting knowledge can facilitate various transfer learning approaches, such as value and policy transfer, without increasing computational complexity. It can be applied across different goals, environments, and RL algorithms, including value-based, policy-based, tabular, and deep RL methods. Experimental evaluations demonstrate the framework's effectiveness in enhancing sample efficiency, especially in challenging main tasks, demonstrated through both a simple Random Walk and more complex optimal control problems with constraints.
\end{abstract}




\end{frontmatter}


\section{Introduction}

Intelligent and data-driven decision-making is becoming an increasingly important part of data science. The subfield of AI concerned with how intelligent agents learn to make optimal decisions in an unknown environment from data and experience is RL. The area has flourished recently by building on advancements in deep learning that have given RL agents superb generalization abilities~\citep{mnih2015}. Deep RL has had a huge impact in multiple areas where it has solved challenging problems, e.g., in game playing \citep{silver2017}, 
financial markets \citep{meng2019}, 
robotic control \citep{ kober2013}, 
optimal control \citep{126844}, 
healthcare \citep{CORONATO2020101964, 10.1145/3477600}, 
autonomous driving \citep{shalev2016},
and recommendation systems~\citep{rec1}. 


A central challenge hindering RL from achieving its full potential in many application areas is that RL approaches generally need a lot of samples to learn good decision policies~\citep{yu2018towards, sung2017learning, beikmohammadi2024accelerating}.  
This translates to costly and time-consuming interactions with the environment, particularly in real-world scenarios like cyber-physical systems, medical applications, and robotics~\citep{NEURIPS2020_8df6a659, NEURIPS2020_448d5eda}. 
The quest to enhance exploration and learning efficiency remains a key challenge in RL.

Most existing RL algorithms start the learning process from scratch. In particular, the agent starts exploring each environment without using any prior knowledge (there are notable exceptions; see Section \ref{section2}). This is fundamentally different from how humans explore and learn in new environments~\citep{GOTTLIEB2013585}. 
When tackling complex tasks, humans adopt a step-by-step approach. They break down the overarching goal into manageable short-term objectives that align with the main goal. To illustrate, consider a math teacher guiding a student. The teacher starts with basic addition, then moves to multiplication, and eventually introduces exponents. This incremental approach enhances comprehension. Similarly, RL agents facing tough challenges might benefit from mastering simpler, related tasks before tackling the primary objective, mirroring human learning principles.

\subsection{Contribution} \label{contrib}
In this paper, we have extended our method originally introduced in \citep{beikmohammadi2023ta}, focusing on the framework for Teacher-Assisted exploration in RL. 
Our framework systematically aids the RL agent's learning process by providing progressively challenging auxiliary goals alongside the main goal. These auxiliary goals are incorporated by introducing an assistant reward in conjunction with the target reward, managed through an annealing function, creating a sequential progression of Markov decision processes (MDPs). The agent alternates between learning each auxiliary goal for a single iteration, leveraging the acquired knowledge to expedite mastery of the main goal. This approach eliminates the need for defining numerous separate MDPs with similar distributions. Importantly, our framework is algorithm-agnostic and compatible with various RL methods, including value-based, policy-based, tabular, and deep RL approaches.

While the core concept was briefly outlined in our prior paper \citep{beikmohammadi2023ta}, here we not only provide a comprehensive exposition of our methodology but also significantly extend our experimental validation.
Specifically, we augment our evaluation by incorporating both linear and non-linear control problems, thus demonstrating the versatility and effectiveness of our approach across varied domains. Furthermore, to ascertain the efficiency of our methodology, we conduct a comparative analysis with state-of-the-art deep RL algorithms (refer to Section \ref{section5}). This comparative evaluation serves to underscore the efficacy of our proposed method in accelerating learning, particularly in challenging tasks, without increasing computational complexity.

Additionally, in this paper, we undertake an extensive literature review, delving deeply into existing efforts aimed at addressing the sample efficiency problem in RL. By contextualizing our work within the broader landscape of RL research, we elucidate the novelty and significance of our contributions, underscoring the advancements made by our framework in ameliorating the sample efficiency challenge.

To facilitate reproducibility and further research in this domain, we have made our code publicly available at \url{https://github.com/AliBeikmohammadi/TA-Explore}.

\subsection{Limitation} \label{limit}
We assume that it's possible to define auxiliary goals and corresponding assistant rewards for certain tasks. While not applicable to every task, surprisingly often, we can create straightforward and 
intuitive assistant rewards to guide the learning process. For instance, in many real-world control problems, the objective is to meet various constraints while optimizing the main goal. Our examples in Section \ref{section5} demonstrate that first learning to satisfy constraints as an auxiliary goal before pursuing the main goal is highly effective in terms of learning speed and sample efficiency. Furthermore, we provide a detailed explanation of the process of defining auxiliary goals and rewards in Section \ref{section4} to illustrate the simplicity of our approach and overcome its limitations.


\section{Our Framework: \texttt{TA-Explore}}
\label{section3}
In this section, we first describe the problem formulation in Section \ref{Sec:RL} and then our novel \texttt{TA-Explore} framework in Section \ref{Sec:TA}.

\subsection{Problem Formulation: MDP}
\label{Sec:RL}
We consider RL in a MDP. Formally, an MDP is characterized by a 5-tuple $(S,A,P,R^T,\gamma)$ where $S$ denotes the set of states; $A$ denotes the set of actions; $P : S \times A \rightarrow \Delta(S)$ denotes the transition probability\footnote{i.e., $P(s'|s,a) = \mathsf{Pr}[s_{t+1} = s' | s_t=s,a_t=a]$.} 
from state $s \in S$ to state $s' \in S$ when the action $a \in A$ is taken; $R^T : S \times A \times S \rightarrow \mathbb{R}$ is the immediate reward\footnote{Here we use the superscript T on the reward to distinguish it from the assistant reward defined later.} 
received by the agent after transitioning from $(s,a)$ to $s'$;  $\gamma \in[0,1)$ is the discount factor that trades off the instantaneous and future rewards. The $\gamma$ controls how much weight is put on old  rewards compared to new ones, i.e., small $\gamma$ means that we put a higher priority on recent rewards. 

We consider episodic tasks, where the episodes are indexed by $e\in \mathbb{N}$. 
 At the beginning of each episode, the agent starts in an initial state $s_0\in S$ which is an IID sample from the distribution $\mu$ on $S$. 
After that, at each time step $t\in \mathbb{N}$, the agent takes action $a_t$ which leads the system to transfer to $s_{t+1} \sim P(\cdot|s_t,a_t)$. The agent receives an instantaneous reward $R^T(s_t,a_t,s_{t+1})$. The agent makes the decision by following a parameterized policy $\pi : S \times \Theta \rightarrow \Delta(A)$, a mapping from the state space $S$ to a distribution over the action space $A$. In particular, we have $a_t \sim \pi(\cdot|s_t; \theta)$ where $\theta \in \Theta$ is an adjustable parameter. The goal of the agent is to find the policy  $\pi(\cdot|s_t; \theta)$ by tuning the parameter $\theta$
that optimizes the cumulative reward and 
to solve the optimization problem
\begin{equation} \label{T}
\small
  \max_{\theta\in \Theta} T(\theta):=\mathbb{E}\bigg[\sum_{t= 0}^{H} \gamma^t R^T(s_t,a_t,s_{t+1})\Big|a_t \sim \pi(.|s_t,\theta),s_0\sim \mu\bigg],
\end{equation}
where $H$ is the termination time.\footnote{Note that $H$ is generally a random variable parametrized by $\theta$.}
We use the notation $T(\theta)$ to indicate that optimizing the above cumulative reward is the main goal of the agent. 


RL aims to find the optimal $\theta$ through trial and error, with the agent constantly engaging the environment. Learning to optimize the main goal can be challenging, often requiring thousands or even millions of interactions to discover effective policies. 
Next, we demonstrate our key concept: enhancing learning efficiency through assistant rewards.

\subsection{\texttt{TA-Explore}}
\label{Sec:TA}
Similar to human learning, the agent should begin by tackling easy tasks resembling the challenging main goal $T(\theta)$. As proficiency grows, progressively intensify the simplicity of these tasks until they align with the main objective.
Formally, we may consider some auxiliary goal
\begin{equation} \label{A}
A(\theta) :=  \mathbb{E}\bigg[\sum_{t= 0}^{H} \gamma^t R^A(s_t,a_t,s_{t+1})\Big|a_t \sim \pi(.|s_t,\theta),s_0\sim \mu \bigg],
\end{equation}
 where $R^A$ is an assistant reward.
Then, the auxiliary goal is to optimize the following problem
\begin{equation*}
\max_{\theta \in \Theta} ~A(\theta).
\end{equation*}
 
Ideally, we should choose $R^A$ in a way that:
\begin{enumerate}[(I)]
    \item it results in a \textit{simple} RL problem that can be solved \textit{fast}. 
    \item solving it is a \textit{side-step towards} the main goal.
\end{enumerate}

Toward achieving the first requirement, let first define $\alpha_{A,T}$, $t_{A}$, and $t_{T}$.
We define $\alpha_{A,T}$ the Pearson correlation coefficient (\( \rho \)) between \( \{R^A(s, a, s')\}_{s \in S, ~a \in A,  ~s' \in S} \) and \( \{R^T(s, a, s')\}_{s \in S, ~a \in A,  ~s' \in S} \), which by the definition of  \( \rho \) ranges in $\alpha_{A,T} \in [-1, 1]$.
Mathematically, this can be expressed as:
\begin{equation} \label{alpha}
\rho(R^A, R^T) := \alpha_{A,T}.
\end{equation}

Also, let \( t_A \) represent the minimum time step such that the \( L_2 \)-norm between the $Q$-function \( Q_{t}^A \) at time \(t = t_A \) for task $A$ and the optimal $Q$-function \( Q^* \) falls below a predefined threshold \( \epsilon \). Formally,
\begin{equation} \label{t_a}
t_A := \min \{ t : \| Q_{t}^A - Q^* \|_2 \leq \epsilon \}
\end{equation}
where the $Q$-function \( Q_t\) computed according to the following recursion:
\begin{equation*} \label{Q}
Q_{t+1}^{} := \left(1-\eta \right)Q_{t}^{}+\eta \left( R + \gamma PV
\right) 
\end{equation*}
where 
\begin{equation*} \label{V}
V(s) := \max _{a^{} \in \mathcal{A}} Q\left(s^{}, a^{}\right)
\end{equation*}
and $\eta \in (0,1)$ is the learning rate.
Moreover, the optimal $Q$-function \( Q^* \) defined as:
\begin{equation*} \label{Q*}
Q^* := \left( R + \gamma PV
\right) 
\end{equation*}

Now, we formally define \textit{Task Similarity} and \textit{Task Simplicity} as follows:
\begin{definition}[Task Similarity] \label{Task_Similarity}
Tasks $A$ and $T$ are considered $\alpha_{A,T}$-similar if 
they satisfy the following conditions:
\begin{enumerate}
    \item Both tasks have the same state space \( S \), action space \( A \), and transition probability $P$.
     \item $\alpha_{A,T} > 0$.
\end{enumerate}
\end{definition}
\begin{definition}[Task Simplicity] \label{Task_Simplicity}
Task $A$ is considered simpler than task $T$ if
they satisfy the following conditions:
\begin{enumerate}
    \item They are \textit{$\alpha_{A,T}$-similar}.
     \item $t_A < t_T$.
\end{enumerate}
\end{definition}

We illustrate examples of such auxiliary tasks
that can truly make learning faster in the next two sections. 

As for satisfying the second requirement, we know that
the agent's primary objective is not mastering these auxiliary goals; rather, it should leverage them to expedite progress towards $T(\theta)$.
To ensure this, the agent should gradually shift its focus towards $T(\theta)$.
We define the parameter $\beta(e)$ to control this transition as follows:
\begin{definition}[$\beta(e)$] \label{beta(e)}
$\beta(e)$ can be considered a subset of any \textit{strictly decreasing} function (i.e., $\beta(e+1) < \beta(e)$) in which $\beta(0) \in (0,1]$ and $\lim_{e\rightarrow\infty} \beta(e) = 0$\footnote{$e\in \mathbb{N}$ represents the episode index.}
\end{definition}
Given $\beta(e)$, now we characterize the new immediate reward $R^e$ as a convex combination of $R^A$ and $R^T$, which the agent should employ during episode $e$ as\footnote{We can also apply this idea on continuous tasks with infinite horizon, but then $\beta(\cdot)$ should be a function of $t$.}:
\begin{align}\label{eqR}
R^e(s_t,a_t,s_{t+1}) \hspace{0.3cm}{:=} \hspace{0.3cm}  \beta(e)R^A(s_t,a_t,s_{t+1}) {+} 
  (1{-}\beta(e))R^T(s_t,a_t,s_{t+1}).
\end{align}
 
 The agent then tries to optimize the following teacher-assisted goal at episode $e$:
\begin{equation} \label{TAe}
     \texttt{TA}^e(\theta)=\mathbb{E}\bigg[\sum_{t= 0}^{H} \gamma^t R^e(s_t,a_t,s_{t+1})\Big|a_t \sim \pi(.|s_t,\theta),s_0\sim \mu \bigg],
\end{equation}
 by finding  the solution to the optimization problem 
 \begin{equation}\label{problem:TA}
\max_{\theta \in \Theta} \texttt{TA}^e(\theta)
\end{equation}
as $e$ goes to infinity. 
When examining $\texttt{TA}^e(\theta)$, we observe that despite having only one assistant reward, multiple teacher-assisted goals are generated due to the presence of the function $\beta(\cdot)$. These goals correspond to several artificial tasks represented as MDPs.

\begin{remark}
\normalfont{
Selecting an appropriate $\beta(e)$ is crucial for smooth learning and influences the number of MDPs. by definition \ref{beta(e)}, $\beta(e)$ should decrease with increasing $e$ and approach 0 for convergence to the main goal. The rate of decrease depends on the task, making it challenging to specify a universal formula. However, the alignment between main and auxiliary goals can guide $\beta(e)$ selection: strong alignment allows for a slower linear decrease, such as $\beta(e) = [(E - e)\beta(0)/E]_+$, starting from $\beta(0)$ and reaching 0 at episode $E$. Conversely, weak alignment suggests a faster exponential decrease, like $\beta(e) = \beta(0)\lambda^e$, with $\lambda \in (0,1)$.}
One method of assessing this alignment is through \textit{Task Similarity} as defined in \ref{Task_Similarity}. The higher the $\alpha_{A,T}$-similarity, indicating a stronger alignment, a gentler decrease is needed (See Sections \ref{section4} and \ref{section5}).
\end{remark}

It is noteworthy that since $\beta(e)$ gradually diminishes to zero, $\theta_{T}^{\star}=\theta_{\texttt{TA}}^{\star}$, where $\theta_{T}^{\star}$ and $\theta_{\texttt{TA}}^{\star}$ are, respectively, the solutions to~\eqref{T} and~\eqref{problem:TA} as $e$ goes to infinity\footnote{Here, we assume a unique optimal solution but, in general, the set of optimizers of~\eqref{T} and~\eqref{problem:TA} are equal.}. This convergence occurs because the influence of the assistant reward diminishes as $e$ increases, solving the \texttt{TA}-goal becomes equivalent to solving the main goal. 
Mathematically, easily, we can show that tasks $\texttt{TA}^e$ and $T$ are:
\begin{enumerate}
    \item $\alpha_{\texttt{TA}^e,T}$-similar.
    \item $\alpha_{A,T} \leq  \alpha_{\texttt{TA}^e,T} \leq \alpha_{T,T}=1$
    \item $\lim_{e\rightarrow\infty} \alpha_{\texttt{TA}^e,T} = 1 $
\end{enumerate}
\begin{proposition} \label{proposition1}
Let tasks $\texttt{TA}^e$, and $T$ are defined by \eqref{TAe}, and \eqref{T}, respectively. Since $R^e$ is a convex combination of $R^A$ and $R^T$ according to \eqref{eqR}, and tasks $A$ and $T$ are $\alpha_{A,T}$-similar by definition \ref{Task_Similarity}, both tasks $\texttt{TA}^e$, and $T$ also have same state space, action space, and transition probability, and $0 \leq \alpha_{A,T} \leq  \alpha_{\texttt{TA}^e,T}$. Then, tasks $\texttt{TA}^e$, and $T$ are  $\alpha_{\texttt{TA}^e,T}$-similar.
\end{proposition}

\begin{proposition} \label{proposition2}
Let tasks $\texttt{TA}^e$, and $T$ are $\alpha_{\texttt{TA}^e,T}$-similar, then according to definition \ref{beta(e)}, $\max \beta(e) = \beta(0) \leq 1$.
Assuming $\beta(0) = 1$, then $R^e = R^A$. Since tasks $A$ and $T$ are $\alpha_{A,T}$-similar by definition \ref{Task_Similarity}, then $\texttt{TA}^e$, and $T$ are at least $\alpha_{A,T}$-similar (i.e., $\alpha_{A,T} \leq  \alpha_{\texttt{TA}^e,T}$).
On the other hand, $\min \beta(e) = 0$, then $R^e = R^T$. Since $\alpha_{T,T} = \rho(R^T, R^T) = 1$, then $\texttt{TA}^e$, and $T$ are at most $1$-similar (i.e., $\alpha_{\texttt{TA}^e,T} \leq  1 $). 
\end{proposition}

More importantly, we can show that task similarity of $\texttt{TA}^e$ and $T$ is strictly increasing until $\alpha_{\texttt{TA}^e,T} = 1 $, meaning the agent shifts toward task $T$ more.

\begin{proposition} \label{proposition3}
Let tasks $\texttt{TA}^e$, and $T$ are $\alpha_{\texttt{TA}^e,T}$-similar, then according to definition \ref{beta(e)}, since $\beta(e)$ is strictly decreasing function, then $1-\beta(e)$ is strictly increasing function. So, considring \eqref{eqR}, the contribution of its first term is strictly decreasing, while the contribution of its second term is strictly increasing. Then $\alpha_{\texttt{TA}^e,T} = \rho(\texttt{TA}^e, R^T)$ is strictly increasing (i.e., $\alpha_{\texttt{TA}^{e+1},T} > \alpha_{\texttt{TA}^e,T}$), if $\alpha_{\texttt{TA}^e,T} \neq 1 $.  
\end{proposition}

In the next sections, we illustrate examples of such strictly increasing behavior of $\alpha_{\texttt{TA}^e,T}$ to corroborate this proposition.



\begin{remark}
\normalfont{Moving to a new MDP in each episode until $\beta(e)=0$ heightens the non-stationarity problem. This aligns with methods resembling exploration encouragement \citep{NIPS2016_afda3322,STREHL20081309}, initially appearing to exacerbate problem instability and slow learning. Nonetheless, it's observed that increased feedback actually accelerates learning. Our approach differs by not advocating blind exploration of all states; instead, we harness transferred knowledge across tasks before deep dives (avoiding overfitting and straying from the main goal). The crucial art lies in setting $\beta(e)$ reliably to transfer suitable prior knowledge via the value function or policy, leading to a more sample-efficient algorithm.}
\end{remark}

\begin{remark}
\normalfont{\texttt{TA-Explore} is versatile, compatible with any RL algorithm, and not limited to specific algorithms like R-MAX \citep{ strehl2009} that rely on tight upper bounds. It can be integrated with any non-($\epsilon$, $\delta$)-PAC algorithm, whether it's value-based, policy-based, tabular, or deep RL. In deep RL, prior knowledge improves initial neural network weightings. It accommodates tasks with multiple goals or constraints (see Section \ref{section5}). Unlike conventional transfer learning methods with pre-training and end-training phases, it offers unified, integrated learning that's simpler and faster. Plus, it does not require pre-solved tasks and easily applies to any environment without added computational complexity.}
\end{remark}

In the following, we illustrate how to use \texttt{TA-Explore} with a simple example in Section \ref{section4}, and then its effectiveness in solving difficult real-world problems is further analyzed in Section \ref{section5}.

\section{Random Walk: An Illustration} \label{section4}

\begin{figure}[t] 
     \centering
     \begin{subfigure}{\linewidth}
         \centering
         \includegraphics[width=0.6\linewidth]{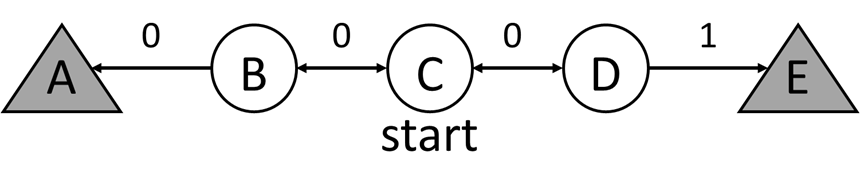}
         \caption{}
         \label{fig1.a}
     \end{subfigure}
     \begin{subfigure}{\linewidth}
         \centering
         \includegraphics[width=0.6\linewidth]{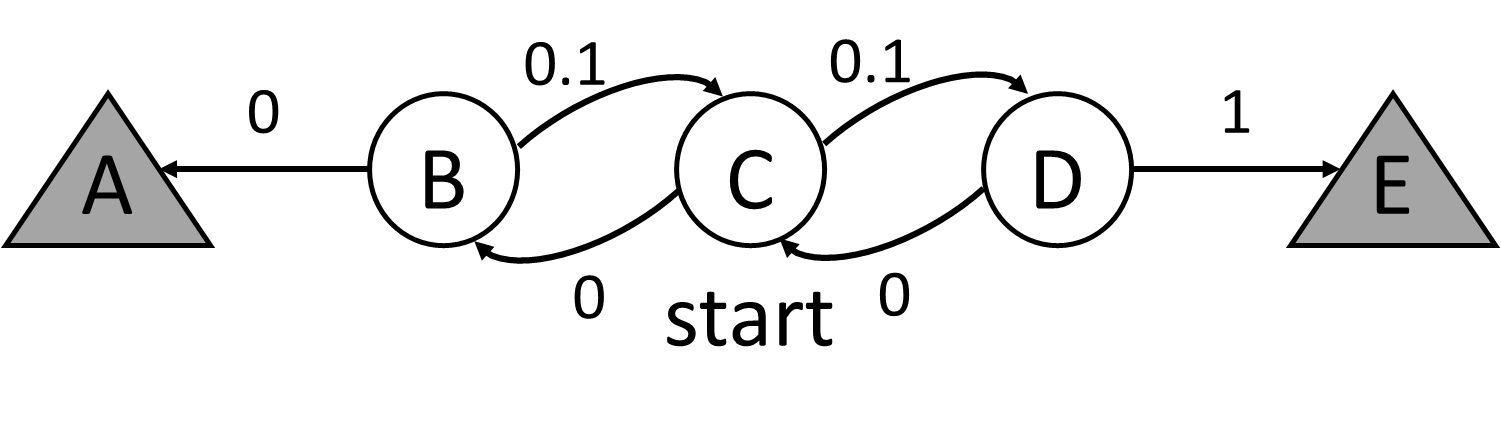}
         \caption{}
         \label{fig1.b}
         \end{subfigure}
        \caption{Random Walk example \citep{sutton2018}, where (a) describes how to receive 
        the target reward 
        $R^T$ and (b) illustrates how to acquire  
        the assistant reward 
        $R^A$. 
        In both cases, the episode terminates by going to A or E states.
        }
        \label{fig1}
\end{figure}

In this section, we showcase the efficiency and simplicity of \texttt{TA-Explore} through an examination of a Random Walk example \citep{sutton2018}, a Markov reward process (i.e., an MDP without actions), allowing us to elucidate the core concepts of \texttt{TA-Explore} in an intuitive manner. 
Figure \ref{fig1} depicts the Random Walk environment comprising 5 states. It encompasses two terminal states, denoted as A and E, alongside three non-terminal states, labeled as B, C, and D. If the agent reaches either terminal state, the episode concludes. In all episodes, the agent initiates from state C. Subsequently, at each time step, the agent moves either to the left or right with an equal probability. The agent receives a reward of zero for every transition, except when it reaches state E on the right, where it garners a reward of $+1$. The discount factor is set to $\gamma = 1$, indicating an undiscounted MDP.
Extending this problem to incorporate additional states is straightforward. We simply arrange all the states in a linear fashion, akin to Figure~\ref{fig1} but with a greater number of states. 
Likewise, the terminal states remain positioned at the ends of the line, both to the left and right. The initial state remains the middle state. 
We will explore random walks with 5, 11, and 33 states.

\begin{figure*}[t] 
    \centering
    \includegraphics[width=0.33\textwidth]{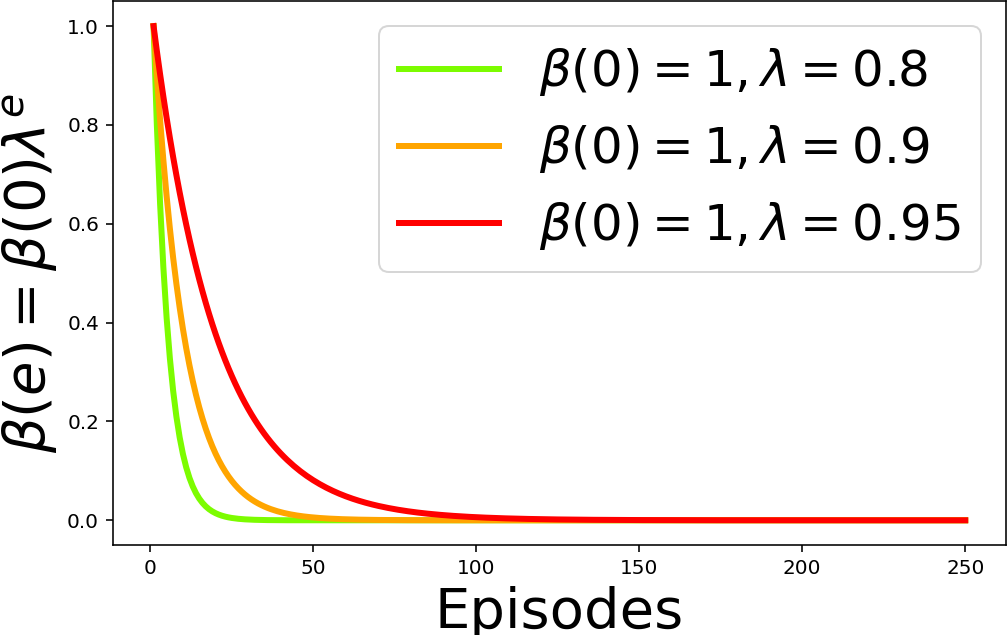}
    \caption{The behaviour of the $\beta(e)$ function considered for Random Walk example, with different $\lambda$ values. 
     As $\lambda$ increases, a slower shifting of the agent from auxiliary goal $A$ learning to main goal $T$ learning happens.}
     \label{fig2}
\end{figure*}

Due to a lack of action here, the agent's main goal is to acquire state values through experience. Initial value learning is slow and necessitates a substantial number of samples. This is because, initially, the agent only receives immediate feedback (a non-zero reward) when it reaches the terminal state on the right. However, we can expedite learning by assigning simpler tasks to the agent that offer more frequent immediate feedback. For instance, in Figure \ref{fig1.b}, we introduce an assistant reward, $R^A$, which provides a reward of 0.1 each time the agent moves to the right, except when it reaches the right terminal state, at which point we award 1. It is evident that learning this auxiliary goal is swifter because the agent receives more frequent reward feedback to update its value function. Consequently, according to the definition \ref{Task_Simplicity}, it can be concluded that task $A$ is simpler compared to task $T$, highlighting $t_A<t_T$.  

\begin{figure*}[t] 
     \centering
      \begin{subfigure}{0.33\linewidth}
         \centering
         \includegraphics[width=\textwidth]{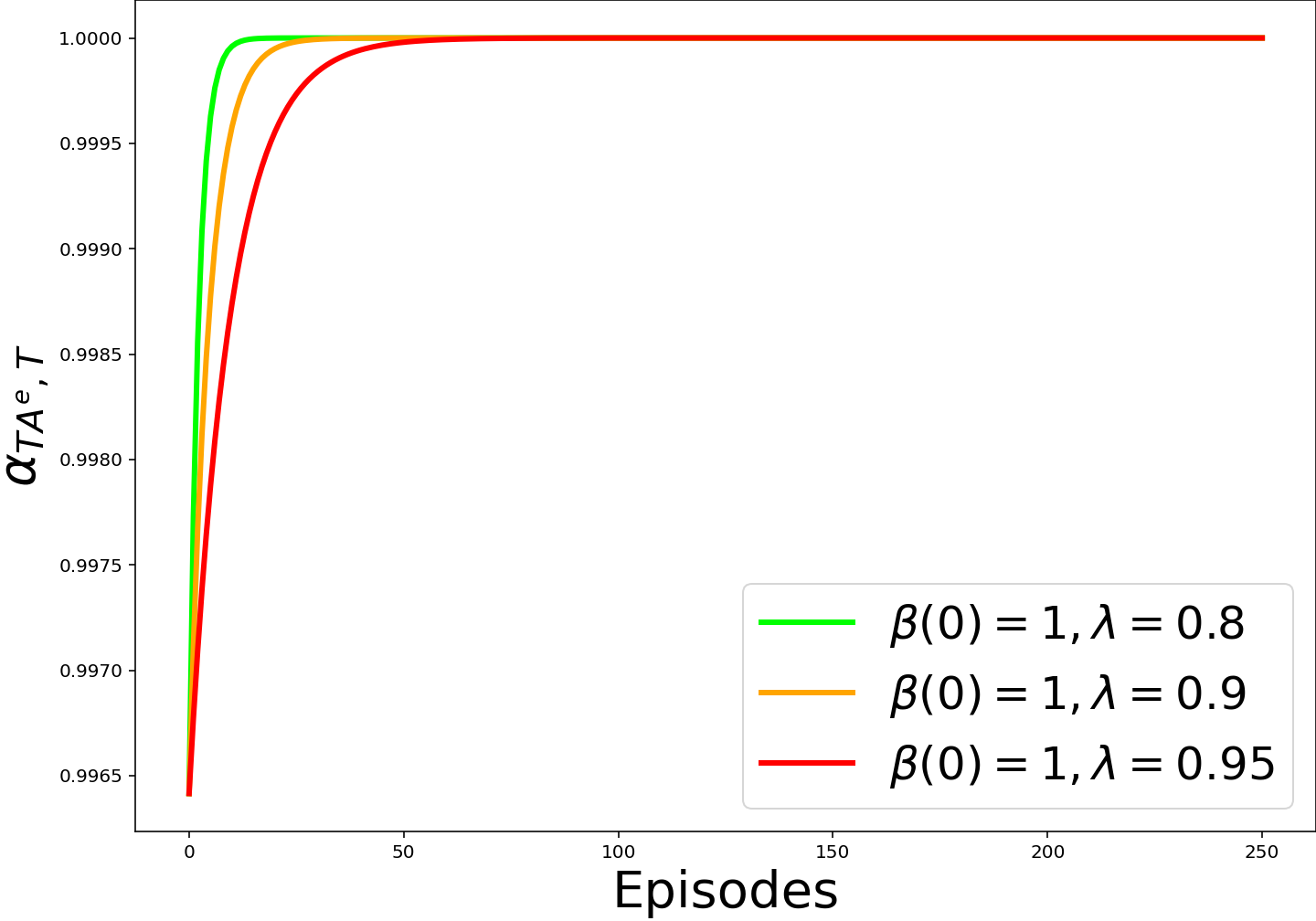}
         \caption{5 states}
         \label{figalpha.a}
         \end{subfigure}
      \begin{subfigure}{0.32\linewidth}
         \centering
         \includegraphics[width=\textwidth]{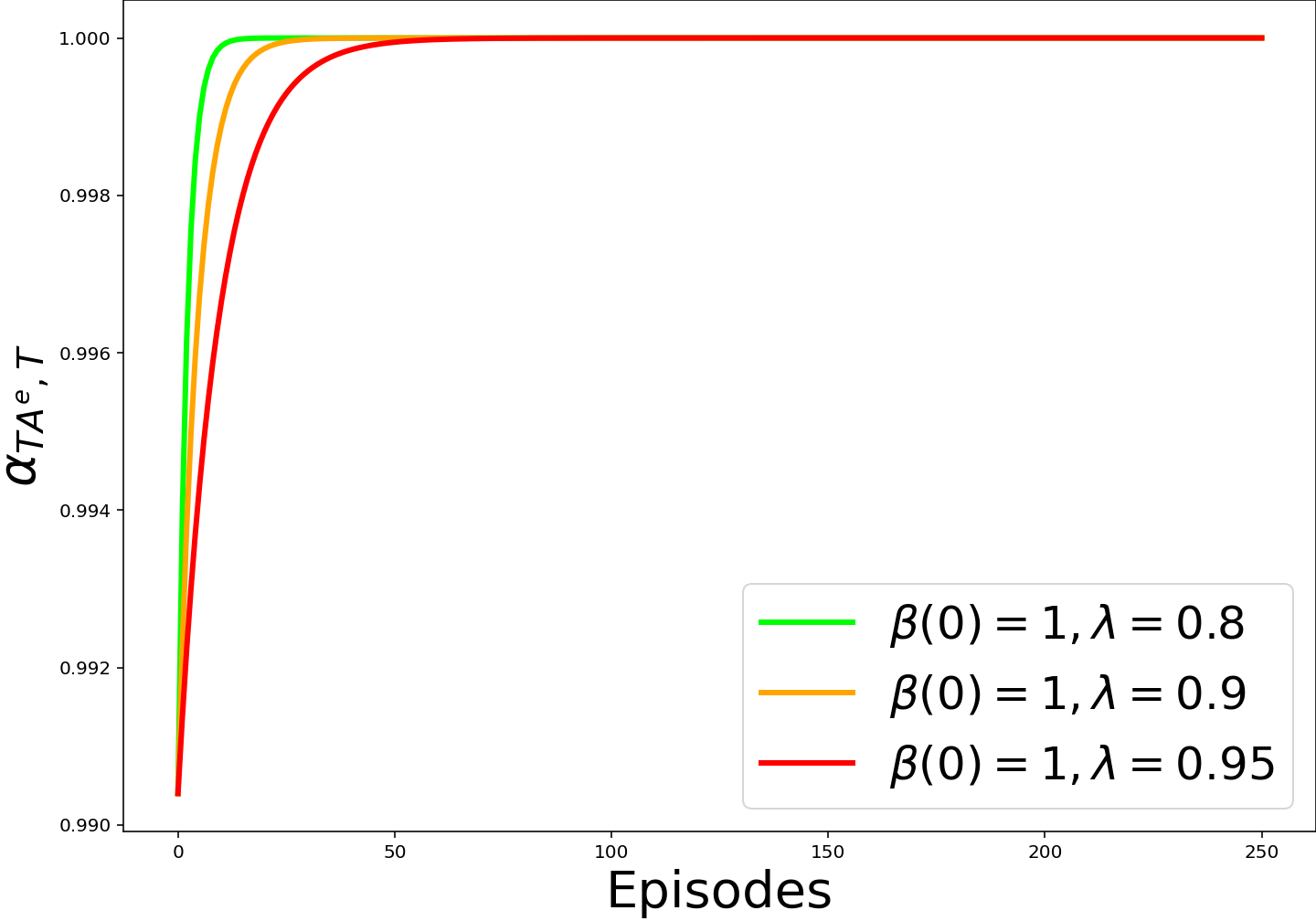}
         \caption{11 states}
         \label{figalpha.b}
         \end{subfigure}
      \begin{subfigure}{0.32\linewidth}
         \centering
         \includegraphics[width=\textwidth]{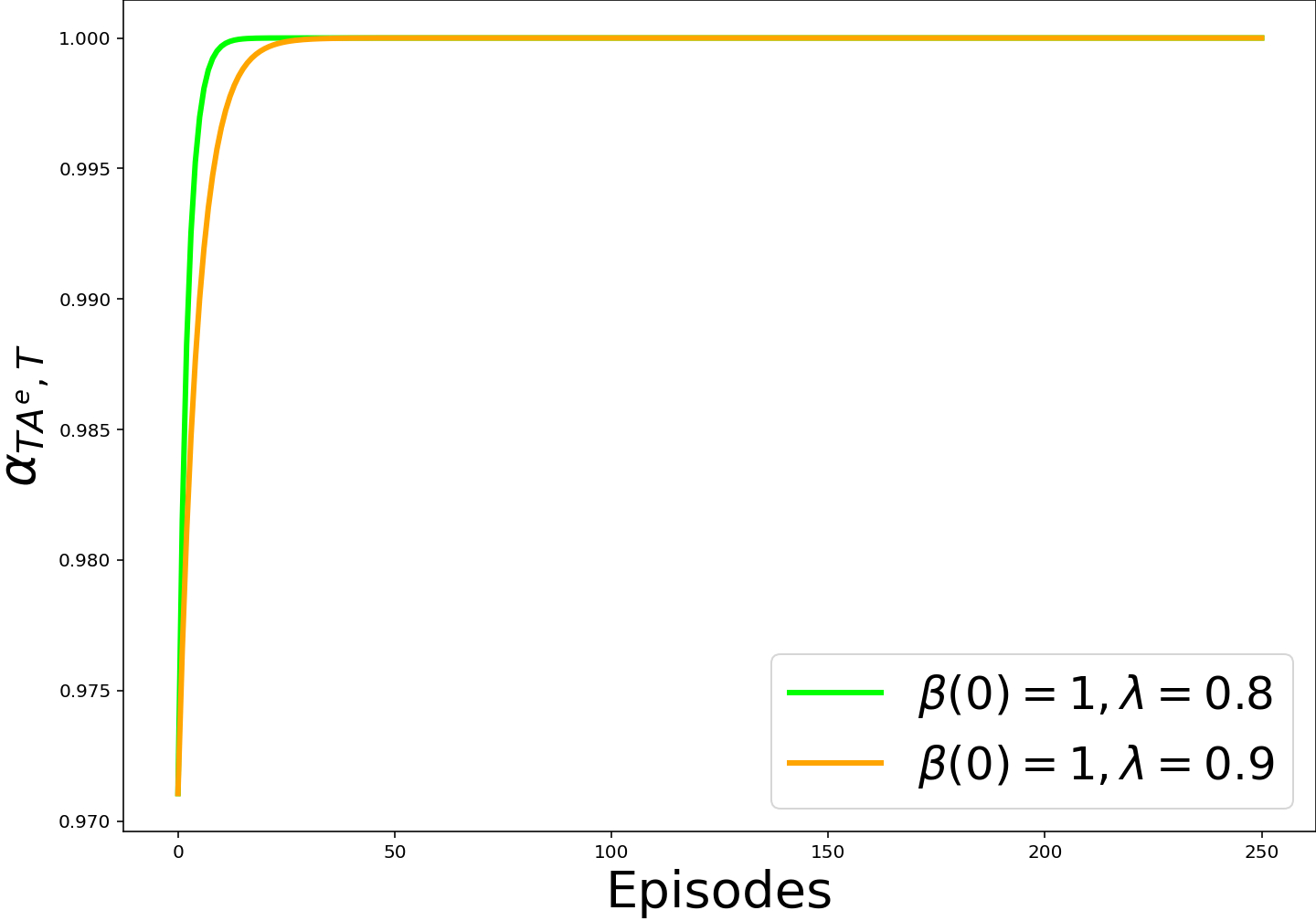}
         \caption{33 states}
         \label{figalpha.c}
         \end{subfigure}
        \caption{The task similarity between $\texttt{TA}$ and the main task $T$ at episode $e$ (i.e., $\alpha_{\texttt{TA}^e,T}$) has been computed for the Random Walk example, considering various numbers of states and $\lambda$ values. 
        Notably, all propositions \ref{proposition1}, \ref{proposition2}, and \ref{proposition3} hold true, demonstrating that $\alpha_{A,T} \leq \alpha_{\texttt{TA}^e,T} \leq 1 $ and $\lim_{e\rightarrow\infty} \alpha_{\texttt{TA}^e,T} = 1 $.}
        \label{figalpha}
\end{figure*}

With the provided $R^A$ as depicted in Figure \ref{fig1.b}, the calculation of $\alpha_{A,T}$ for 5, 11, and 33 states yields values of 0.9964, 0.9904, and 0.9711, respectively, satisfying the conditions in definitions \ref{Task_Similarity} and \ref{Task_Simplicity}.
Then, as depicted in Figure \ref{fig2}, a straightforward yet intelligent approach for determining the $\beta(e)$ function might be $\beta(e) = \beta(0)\lambda^{e}$,  where $\beta(0)=1$, and $\lambda \in \{0.8, 0.9, 0.95\}$. The agent could initially begin learning goal $A$ but will quickly shift its focus to mastering goal $T$.
Figure \ref{figalpha} provides evidence of such shifting by highlighting the growth in the task similarity between $\texttt{TA}$ and the main task $T$ at episode $e$.

To assess the performance of our proposed method, we opted for a basic value-based algorithm, specifically the temporal-difference learning method (TD(0))  \citep{tesauro1995} with a constant step size of 0.1, despite the compatibility of \texttt{TA-Explore} with various RL algorithms. This choice is due to Random Walk's nature as an MRP problem with a discrete state space. 
In this instance, transfer learning occurs through value transfer, where the value functions derived for each task serve as the initial values for the subsequent task. Our results are an average of 100 test runs, with each execution initializing the state-value functions for all states to zero.
The performance measurement hinges on the root mean square (RMS) error between the learned value function and the true value function corresponding to $T$ and $R^T$. 

\begin{figure*}[t] 
     \centering
      \begin{subfigure}{0.33\linewidth}
         \centering
         \includegraphics[width=\textwidth]{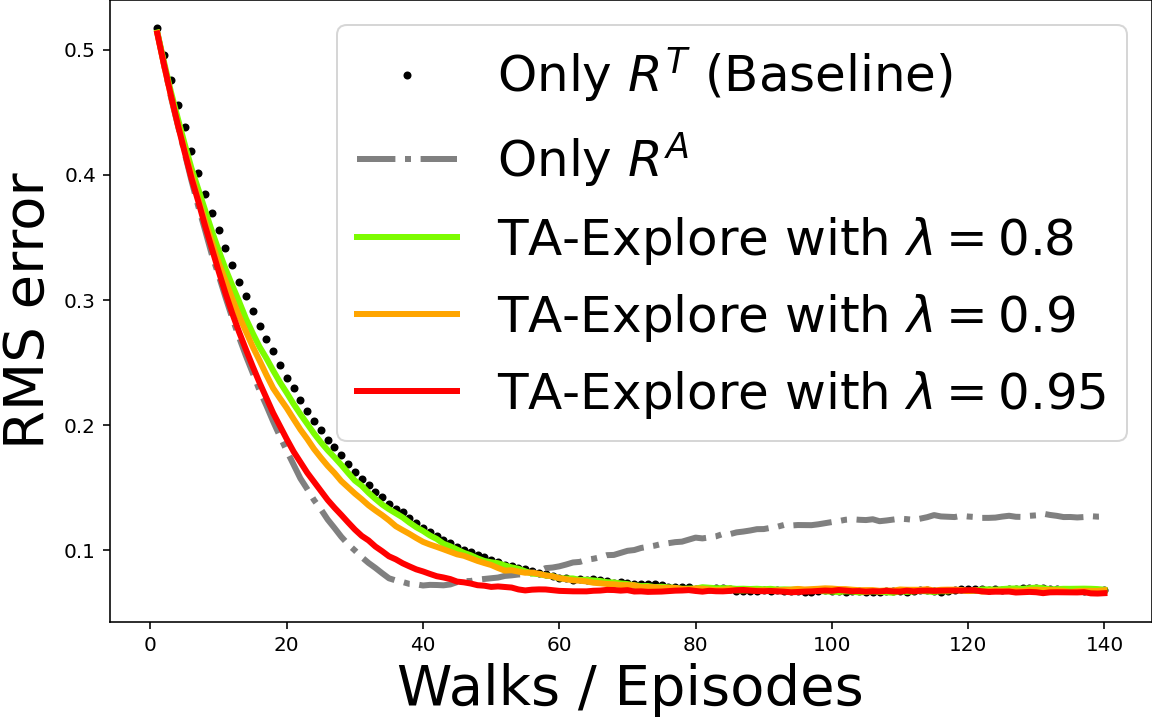}
         \caption{5 states}
         \label{fig3.a}
         \end{subfigure}
      \begin{subfigure}{0.32\linewidth}
         \centering
         \includegraphics[width=\textwidth]{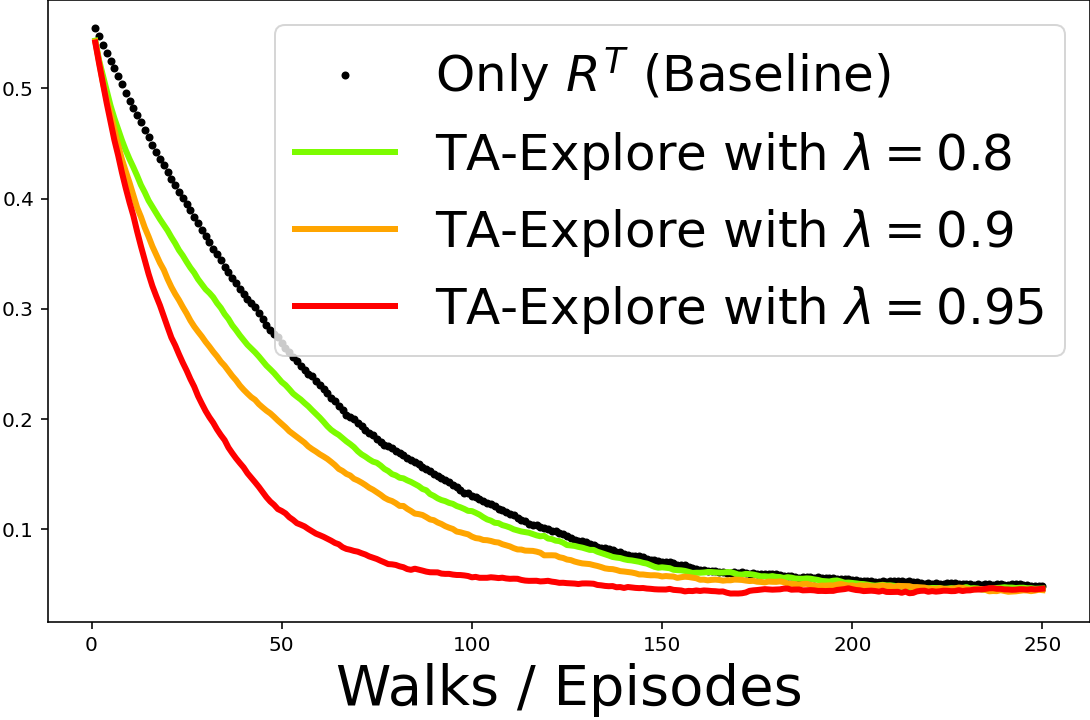}
         \caption{11 states}
         \label{fig3.b}
         \end{subfigure}
      \begin{subfigure}{0.32\linewidth}
         \centering
         \includegraphics[width=\textwidth]{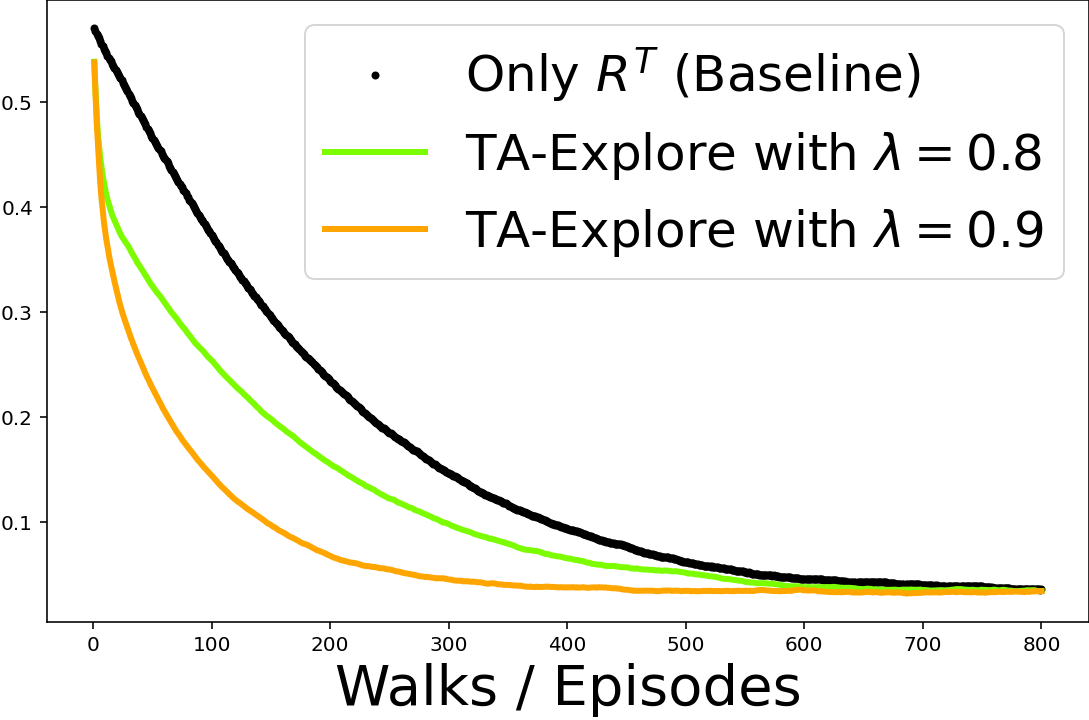}
         \caption{33 states}
         \label{fig3.c}
         \end{subfigure}
        \caption{
        The main goal learning curves in the Random Walk example for different $\lambda$ values and the different number of states. The performance measure shown is the RMS error between the value function learned considering the assumed rewards (i.e., only $R^T$, only $R^A$, or \texttt{TA-Explore} with different $\lambda$s) and the true value function, which is averaged over the states and then averaged over 100 runs. 
        As it turns out, in all cases, by choosing the appropriate $\lambda$, one can be sure that \texttt{TA-Explore} is learning faster. 
        The undrawn curves in (b) and (c) have diverged and have been omitted to show the rest of the results in more detail.}
        \label{fig3}
\end{figure*}

In Figure \ref{fig3.a}, learning $A$ initially yields positive results, with the agent also unintentionally learning $T$ by episode 38, resulting in a reduction in RMS error. However, as the agent prioritizes learning goal $A$, the error increases, reflecting the divergence in the value functions of the two goals; learning $A$ outpaces learning $T$ (refer to Only $R^T$ in Figure \ref{fig3.a}). Therefore, starting with goal $A$ and subsequently transitioning to goal $T$ before overfitting can maximize prior knowledge utilization. This approach accelerates learning $T$. Specifically, with $\lambda = 0.95$, \texttt{TA-Explore} achieves twice faster convergence on goal $T$.

In Figure \ref{fig3.b}, even with 11 states, $\lambda = 0.95$ remains an effective choice. The critical insight is that as the state count rises, our proposed method's superiority becomes evident. More states decrease the likelihood of initially reaching the far-right state, slowing down baseline convergence, which heavily relies on state numbers. Conversely, our method maintains a $50\%$ chance of receiving rewards from the first step. In Figure \ref{fig3.c}, we consider 33 states where the baseline fails to converge, while \texttt{TA-Explore} succeeds. Notably, $\lambda = 0.95$ ceases to be an appropriate choice. 
As the number of states grows, disparities in true state-value functions emerge, even though the two goals initially align. This can be seen also from the value of $\alpha_{A,T}$. 
In such a case, spending too much time on auxiliary goal learning hampers the agent's access to valuable prior information and can lead to extended main goal learning, necessitating time to forget undesirable experiences. Therefore, the agent should transition to main goal learning more quickly, albeit not hastily, as illustrated in Figure \ref{fig3.c} with $\lambda = 0.9$, resulting in more than double the convergence speed.
 

It is reasonable to anticipate that due to its profitability on simple tasks and value-based algorithms that offer limited maneuvering, \texttt{TA-Explore} will demonstrate remarkable performance on more complex tasks, including those with continuous state and action spaces. Furthermore, it is expected to significantly enhance the results of policy-based approaches as well as deep RL methods. The next section will provide evidence supporting this claim.

\section{Experiment on Control Problems} \label{section5}
 
 We illustrate the promise of the \texttt{TA-Explore} framework on two optimal control problems with constraints. We consider a dynamical system governed by the difference equation $s_{t+1}=g_t(s_t,a_t,e_t)$, where $s_t$ is the state of the system, $a_t$ is the control action, and $e_t$ is a random disturbance; $g_t$ is the rule that maps the current state, control action, and disturbance at time $t$ to a new state. The goal of the agent is to find the actions that optimize the accumulated immediate rewards  $f_t(s_t,a_t)$. Often there are also constraints on the states. Thus, the agent should not only optimize its rewards, but it should also ensure the feasibility of the states. This could be, for example, to ensure the safety of the system operations. Mathematically, such problems are  formulated as follows:
\begin{displaymath}
\begin{split}\label{eq:10}
    \text{maximize} & \sum_{t=0}^{H} \gamma^t f_t(s_t,a_t) \\
    \text{subject to } & s_{t+1}=g_t(s_t,a_t,e_t), \text{and }\\   
   & s_{\min} \leqslant s_t \leqslant s_{\max}.
\end{split}
\end{displaymath}
The constraints on state variable $s_t$ pose challenges for optimal control, resembling real-world issues. Even with known linear dynamics $g_t(s_t,a_t,e_t)$ and quadratic rewards, traditional LQ-regulators cannot handle it due to the state constraint~\citep{aastrom2007feedback}. Furthermore, the problem is exacerbated by the typical lack of knowledge about the system's dynamics~\citep{recht2019}. In such cases, learning-based methods, such as RL, are often the only viable solutions.

In RL, constraints are typically incorporated into the reward as the main goal $T$ via a penalty for violations. This means the RL agent must simultaneously learn to satisfy constraints while optimizing rewards. Learning to satisfy constraints is relatively easier because the agent has more action choices, facilitating constraint satisfaction. Hence, it makes sense to use constraint satisfaction as the auxiliary goal $A$ in our framework. We achieve this by employing a negative assistant reward $R^A$. As $R^A$ is part of the target reward $R^T$, there is a strong alignment between them, resulting high $\alpha_{A,T}$-similarity. Consequently, the $\beta(e)$ function can be selected more easily. Our choice is a descending linear function, starting at $\beta(0)$ and reaching zero at $e=E$\footnote{i.e., $\beta(e) = [(E-e)\beta(0)/E]_+$}. This results in a sequence of $E$ MDPs, each solved over one episode\footnote{The source code is  available at \url{https://github.com/AliBeikmohammadi/TA-Explore}}.

\subsection{Optimal Temperature Control with Constraint - Linear Dynamics} \label{section5.1}
Consider a data center cooling where three heat sources are coupled to their own cooling devices.
Each component of the state $s$ is the internal temperature of one heat source, which should be maintained in the range of -2 to 2 (i.e., we have constraints on states such that $-2 \leqslant s_t \leqslant 2$).
Under constant load, the sources heat up and radiate heat into the surrounding environment.  
The voltage of each cooler is known as $a$, and the objective is to minimize it while satisfying the constraint. It can be approximated by the linear dynamical system $s_{t+1}=As_t+Ba_t+e_t$ where,
\begin{displaymath}
A=
  \begin{bmatrix}
1.01 & 0.01 & 0\\
0.01 & 1.01 & 0.01\\
0 & 0.01 & 1.01
\end{bmatrix}~~~~
, ~~~~ B=I,
\end{displaymath}
and where $e_t$ is a noise with zero mean with covariance $10^{-4}I$. The voltage of each cooler is the action and the goal is to minimize the power (i.e., $\omega||a||^2$) while satisfying the constraint \citep{recht2019}. To put it in our framework, we define the target reward $R^T$  and the assistant reward $R^A$  as follows:
\begin{equation} \label{rt}
\begin{split}
  &R^T =
    \begin{cases}
      -\omega \Vert a\Vert ^2 & \text{if constraint is satisfied}\\
      -\omega \Vert a\Vert ^2 -100 & \text{otherwise}
    \end{cases} 
    \\
   &R^A =
    \begin{cases}
      0 & \hspace{1.3cm}\text{if constraint is satisfied}\\
      -100 & \hspace{1.3cm}\text{otherwise}
    \end{cases}  
\end{split}
\end{equation}
where $\omega$ is the weight of the first term; its impact will be examined later.

\begin{figure}[t] 
     \centering
     \begin{subfigure}{0.49\linewidth}
         \centering
         \includegraphics[width=\linewidth]{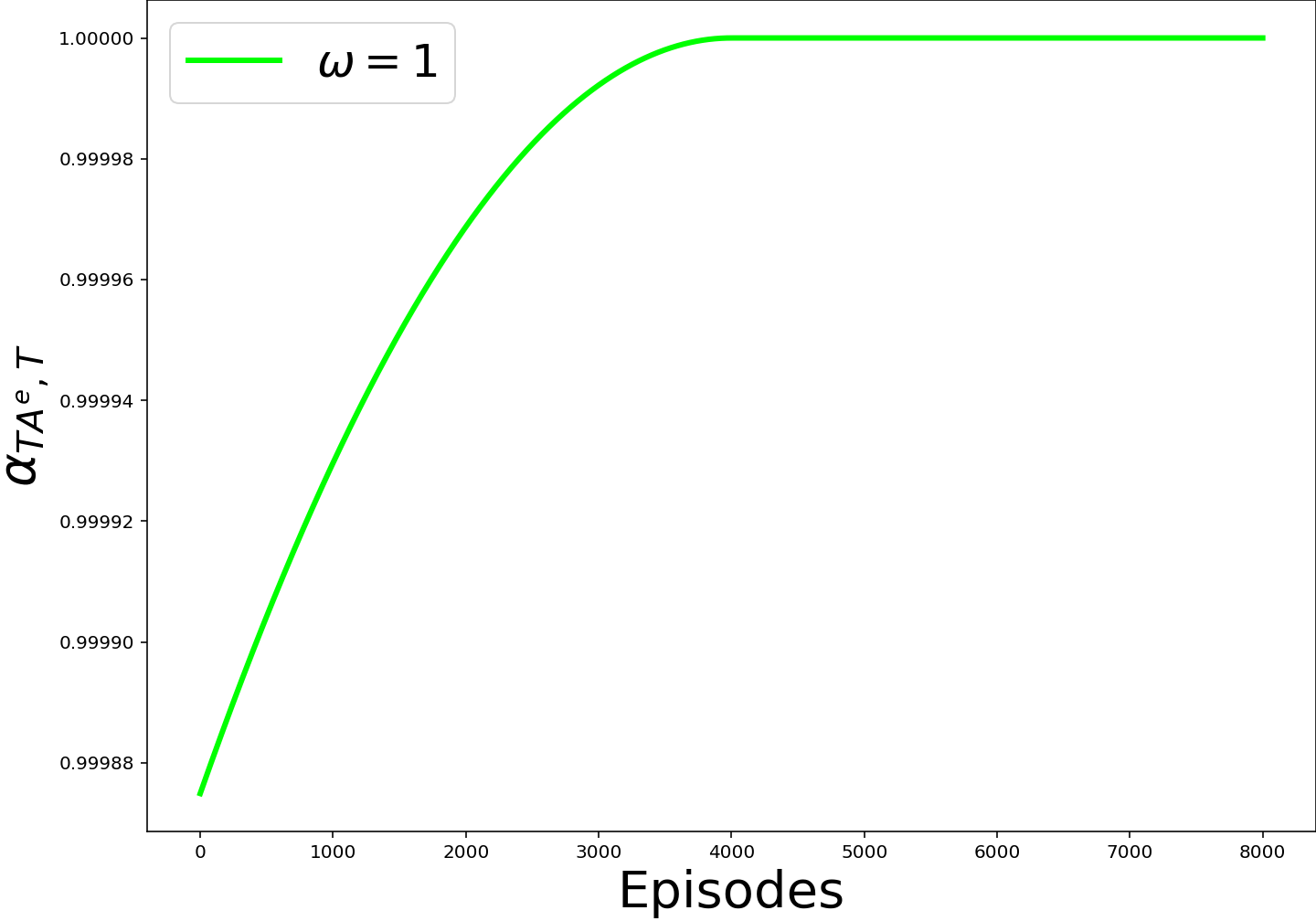}
          \caption{}
         \label{alphaOTCP.a}
         \end{subfigure}
         \hfill
     \begin{subfigure}{0.49\linewidth}
         \centering
         \includegraphics[width=\linewidth]{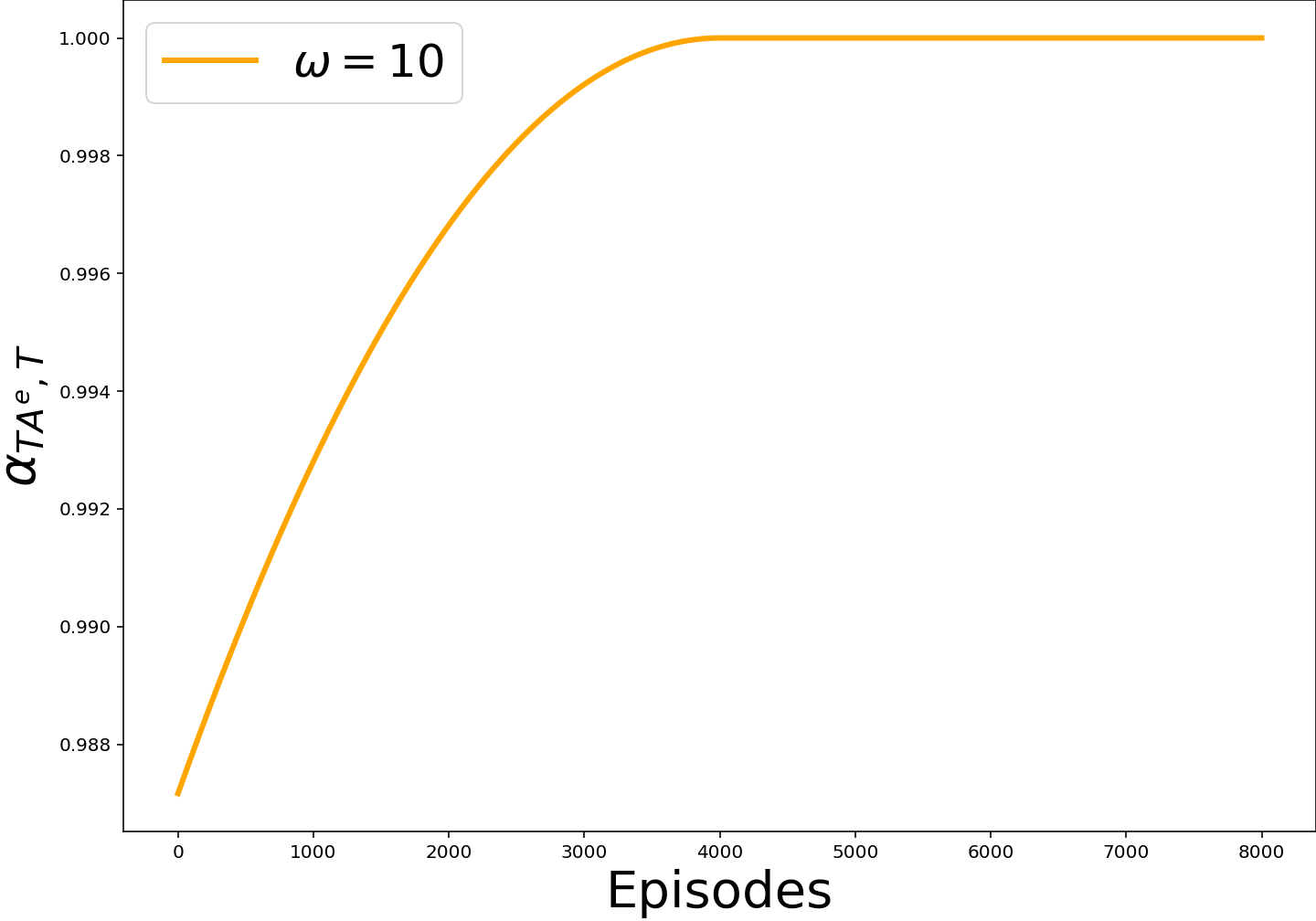}
          \caption{}
         \label{alphaOTCP.b}
         \end{subfigure}
         \hfill
     \begin{subfigure}{0.49\linewidth}
         \centering
         \includegraphics[width=\linewidth]{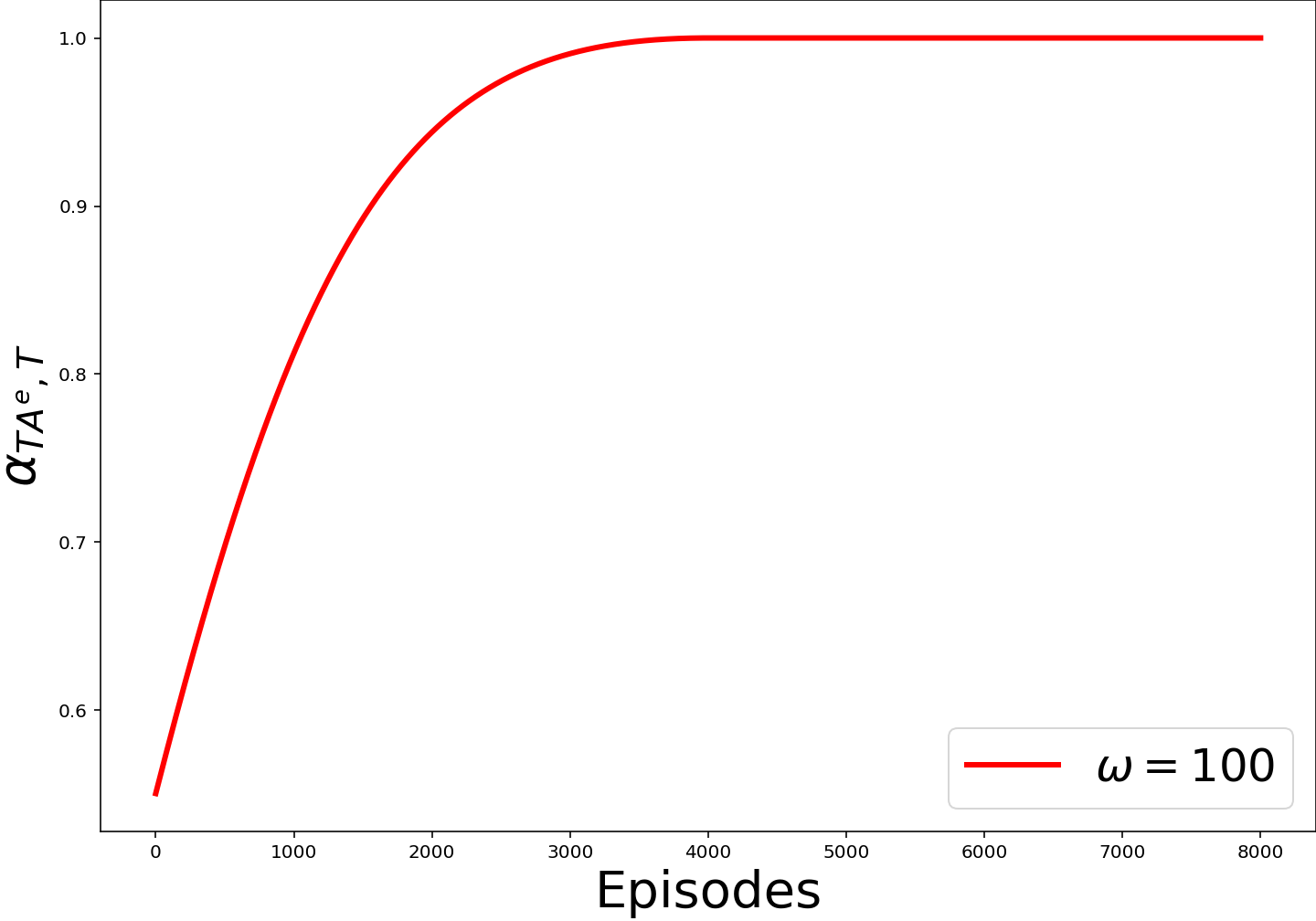}
          \caption{}
         \label{alphaOTCP.c}
         \end{subfigure}
        \caption{The task similarity between $\texttt{TA}$ and the main task $T$ at episode $e$ (i.e., $\alpha_{\texttt{TA}^e,T}$) has been computed for optimal temperate control problem with constraint, considering different weighting values to the control objective (i.e., (a) $1\Vert a\Vert ^2$, (b) $10\Vert a\Vert ^2$, (c) $100\Vert a\Vert ^2$). 
        }
        \label{alphaOTCP}
\end{figure}

The time horizon is 100 steps, and there are 8000 training episodes. To increase the challenge, states can have any initial value following a standard normal distribution. Episodes do not end if the constraint is not satisfied (training continues for 100 time steps in each episode). For $\beta(e)$, we set $E=4000$ and $\beta(0)=1$, so the objective function equals the main goal $T$ after 4,000 episodes.

To train the agent, we employ PPO, a deep RL approach, due to its success as a policy-based algorithm \citep{schulman2017proximal}. PPO is known for its efficiency, presenting a challenge for improvement. In our setup, we update actor and critic models with mini-batches of 512 samples, using 10 training epochs and the Adam optimizer (LR = 0.00025). We set discount and lambda factors at 0.99 and 0.9, respectively.
Both actor and critic models consist of three layers with ReLU activation, with unit sizes of 512, 256, and 64. Actor and critic output layers have 3 and 1 neurons, respectively, utilizing tanh and identity activation functions.
In addition to the original PPO, we compare our performance against four state-of-the-art algorithms: A2C \citep{mnih2016asynchronous}, DDPG \citep{lillicrap2015}, SAC \citep{haarnoja2018soft}, and TD3 \citep{td3}.

\begin{figure}[t] 
     \centering
     \begin{subfigure}{0.49\linewidth}
         \centering
         \includegraphics[width=\linewidth]{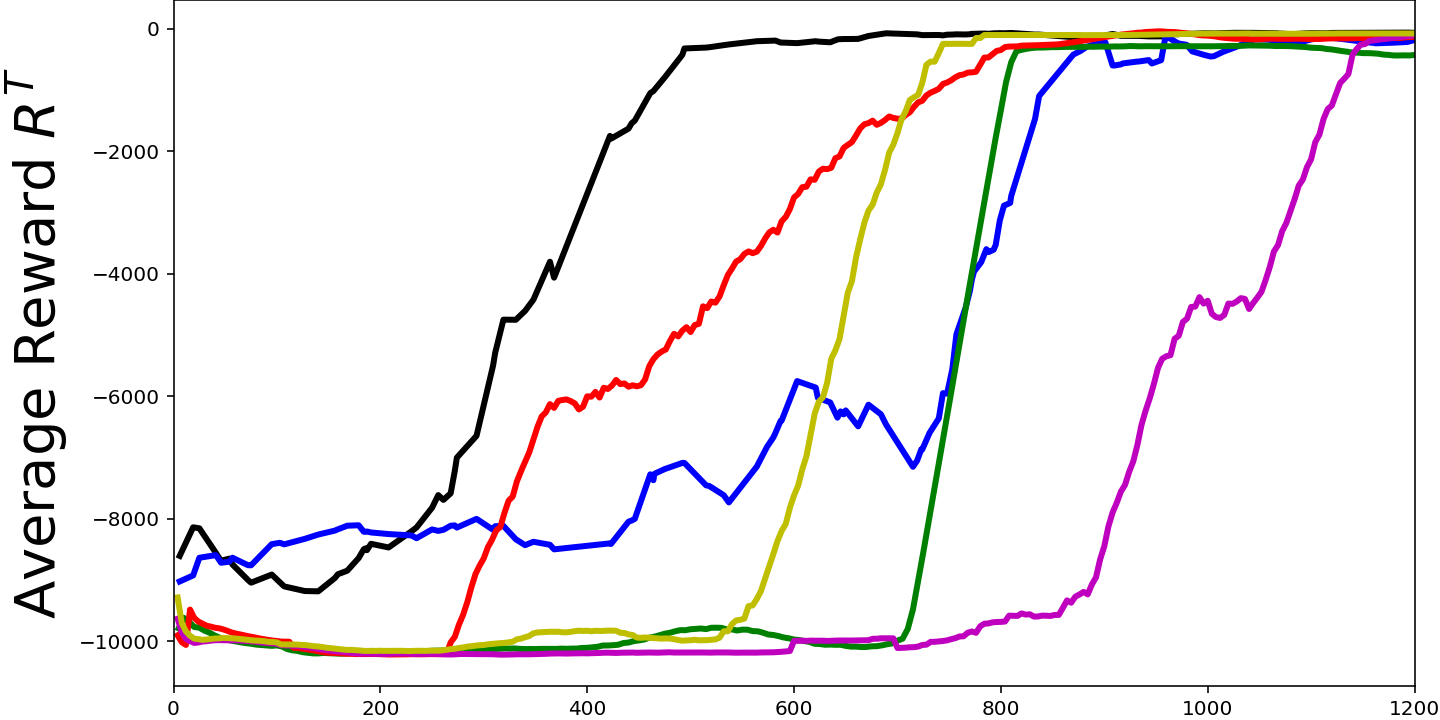}
          \caption{}
         \label{fig4.a}
         \end{subfigure}
         \hfill
     \begin{subfigure}{0.49\linewidth}
         \centering
         \includegraphics[width=\linewidth]{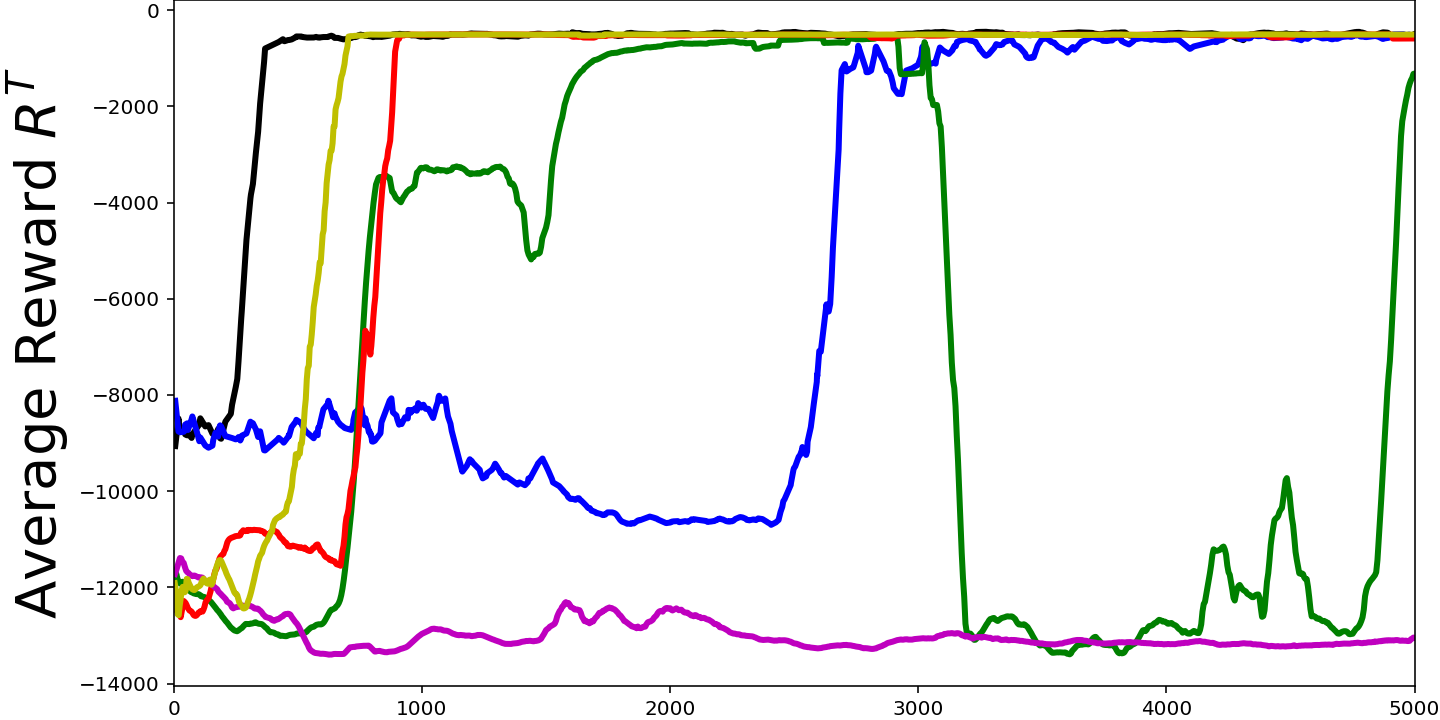}
          \caption{}
         \label{fig4.b}
         \end{subfigure}
         \hfill
     \begin{subfigure}{0.49\linewidth}
         \centering
         \includegraphics[width=\linewidth]{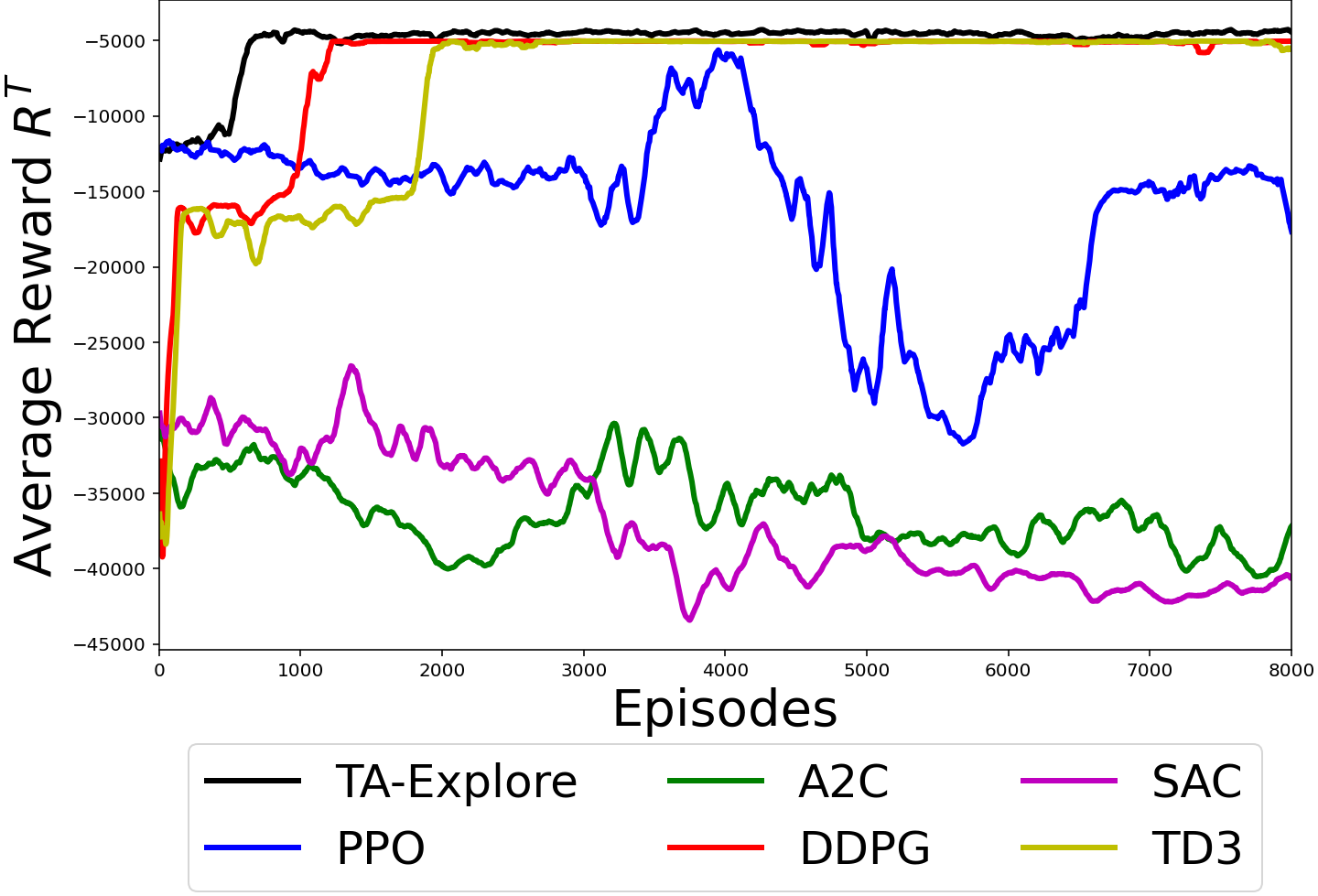}
          \caption{}
         \label{fig4.c}
         \end{subfigure}
       \caption{Average performance of \texttt{TA-Explore} on optimal temperate control with constraint for different weighting values to the control objective (i.e., (a) $1\Vert a\Vert ^2$ , (b) $10\Vert a\Vert ^2$, (c) $100\Vert a\Vert ^2$). The performance measure shown is the amount of the target reward $R^T$ achieved by the agent in each episode, regardless of whether the agent is learning according to $R^T$ or a combination of $R^T$ with the assistant reward $R^A$ according to Equation \eqref{eqR}. 
       To have a clear illustration, 50 episodes moving average reward are plotted. As it turns out, \texttt{TA-Explore}, in all cases, converges very fast. But, the baselines converge too late and in some cases do not converge during 8000 episodes.}
        \label{fig4}
\end{figure}

In this example, transfer learning is achieved via policy transfer using episode-obtained weights for each task as initial neural network weights for the next task. Figure \ref{fig4} displays \texttt{TA-Explore} framework performance. Our framework maintains the same hyperparameters as PPO, streamlining its applicability across various algorithms without requiring further tuning. Other algorithm implementations are based on the Stable Baselines3 repository \citep{stable-baselines3}. Performance evaluation involves plotting the average reward $R^T$, a pertinent metric as the primary focus of the agent is mastering the main goal $T$; the auxiliary goal $A$ serves solely as a learning facilitator. Consequently, acquiring proficiency in the auxiliary goal $A$ has no intrinsic importance, unless it expedites knowledge transfer for faster convergence.

In Figure \ref{fig4.a}, \texttt{TA-Explore} converges twice as fast because of rapid learning of assistant reward $R^A$ in early episodes. Other algorithms exhibit slower convergence due to the agent's struggle in discerning complex rewards, making it challenging to differentiate constraint violations from deviations from the main goal. 

Prioritizing the initial term of $R^T$ sacrifices more information from $R^A$, diminishing alignment between them, thus heightening the challenge. This effect is evident in Figure \ref{alphaOTCP}, where, for example, $\alpha_{A,T}$ values are 0.9998, 0.9871, and 0.5505 for $\omega$ values of 1, 10, and 100, respectively.
Nonetheless, as depicted in Figures \ref{fig4.b} and \ref{fig4.c}, \texttt{TA-Explore} converges nearly as swiftly. In contrast, the baselines yield unsatisfactory results, often failing to converge even after 8,000 episodes. Thus, it is evident that for more intricate problems, our convergence outpaces the baselines. It is worth noting that, while we have demonstrated the significance of learning constraints and gradually incorporating the primary problem, further enhancement can be achieved through skillful formulation of auxiliary goals informed by domain knowledge.

\subsection{Coupled Four Tank MIMO System - Nonlinear Dynamics}

\begin{figure}[ht]
  \centering
  \includegraphics[width=0.65\linewidth]{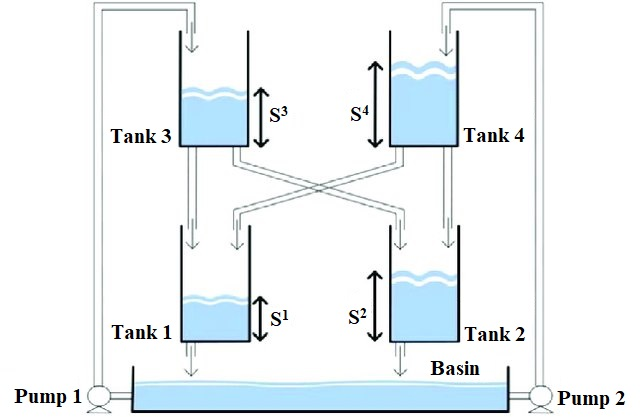}
  \caption{Coupled four tank MIMO system \citep{GOUTA2017280}
  ; known as a nonlinear optimal control problem with constraints
  .}
  \label{fig5}
\end{figure}

We also study the performance on a coupled four-water tank system with nonlinear dynamics \citep{GOUTA2017280}.
Figure \ref{fig5} illustrates how this system is comprised of a liquid basin, two pumps, and four tanks with the same area with orifices. The dynamics of the systems are given by:
\begin{align*}
\begin{cases}
s_{t+1}^1 =&-c_1\sqrt{s_t^1}+c_2\sqrt{s_t^3}+c_3\sqrt{s_t^4} \\
s_{t+1}^2 =&-c_4\sqrt{s_t^2}+c_5\sqrt{s_t^3}+c_6\sqrt{s_t^4} \\
s_{t+1}^3 =& -c_7\sqrt{s_t^3}+c_8 a_t^1 \\
s_{t+1}^4 =& -c_9\sqrt{s_t^4}+c_{10} a_t^2
\end{cases}   
\end{align*}
where $c_1, ..., c_{10}$ are model parameters that are set as in \citep{GOUTA2017280}. The actions are $a^1$ and $a^2$, which indicate  the voltages applied, respectively, to Pump 1 and Pump 2, which are  bounded between 0 and 12Volts. The states $s^1$, $s^2$, $s^3$, and $s^4$ indicate  the liquid levels in the four tanks. They should always be greater than 3cm and less than 30cm (i.e., we have constraints both on states and actions such that $3 \leqslant s_t \leqslant 30$,  $0 \leqslant a_t \leqslant 12$).

The problem can be integrated into our proposed framework using the same rewards from Equation \ref{rt} ($\omega =1$). We employ \texttt{TA-Explore} with $\beta(e) = [(E-e)\beta(0)/E]_+$, where $\beta(0)=0.5$ and $E = 3000$. Initial state values are uniformly distributed between 3 and 30. Episodes are terminated if constraints are not met. All other settings remain consistent with Section \ref{section5.1}, with training requiring 30,000 episodes due to its complexity.
In this experiment, we use PPO with settings identical to those in Section \ref{section5.1}, except the number of actor model output neurons is 2. Ensuring action feasibility is straightforward because the actor model's output, governed by the tanh activation function, scales between 0 and 12.

\begin{figure}[t]
  \centering
  \includegraphics[width=0.6\linewidth]{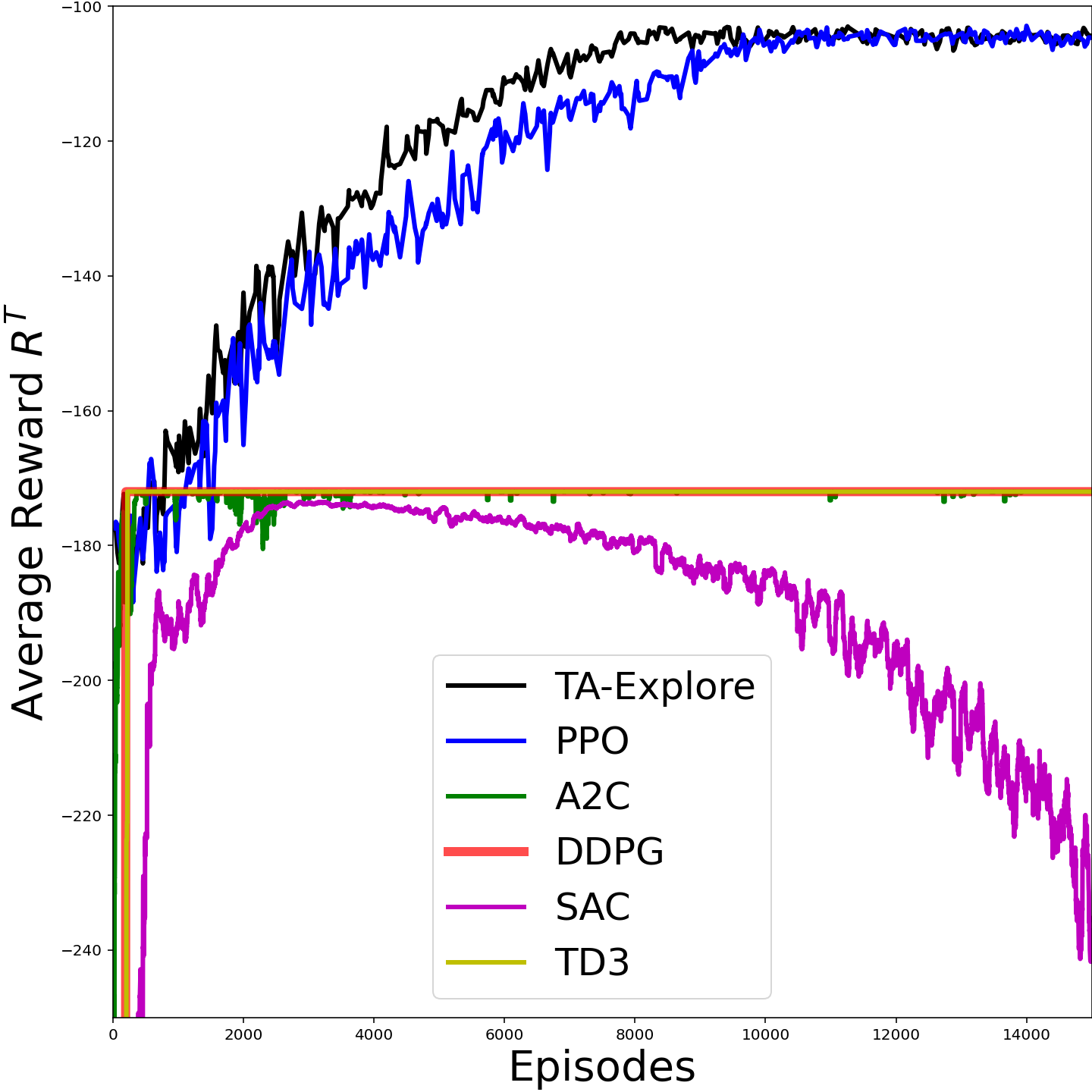}
  \caption{Average performance of \texttt{TA-Explore} on coupled four tank MIMO system. 
  The performance measure shown is the amount of the target reward $R^T$ achieved by the agent in each episode. 
  Note that, 50 episodes moving average reward are plotted to have a clear illustration.}
  \label{fig6}
\end{figure}

Figure \ref{fig6} demonstrates the performance contrast between our proposed approach and baseline methods. Notably, \texttt{TA-Explore} consistently converges to the target reward over $30\%$ faster than PPO, regardless of problem complexity or suboptimal $\beta(e)$ choices.
This remarkable convergence stands in contrast to the other baselines, which struggle to reach convergence. 
The choice of $\beta(0)=0.5$ signifies that from the very first episode, our method prioritizes minimizing voltage while respecting constraints, albeit with half the weight compared to the target reward $R^T$.

\section{Related Work} \label{section2}
To the best of our knowledge, this paper stands out from the rest of the studies because first, we use a sequence of auxiliary goals (i.e., MDPs) by defining only one assistant reward (not many and not following from a particular distribution) and an annealing function. Second, we learn each auxiliary task only one iteration (without necessarily having $\epsilon$-accuracy with $\delta$ probability assumption). And third, our proposed method supports policy and value transfers, which leads our approach to combine with any type of RL-based algorithm to improve sample efficiency. However, the scope of related work can be broadened to include other efforts to help the agent learn its main task more efficiently.
It should be noted that due to the strong assumptions of other studies stated in this section, it is not possible to use them in our experiments. That is why we have to limit the comparison of our method with the original algorithms.

 Improving the exploration process of an RL agent has been well-studied in the literature. 
Many previous efforts are based on the hypothesis that if we can visit as many states as possible with high probability, then the agent's learning will be more successful and sample efficient. In other words, further exploration is in line with achieving the main goal of the RL agent. There are two fundamental questions that emerge in this setting. In the absence of rewards, what are the agents supposed to look for? And at what point should the agent stop exploring and start acting greedily?  
Research that seeks to answer these questions can be divided into two general categories.
Firstly, there are single-environment approaches, which include intrinsic motivation with \textit{visitation counts} \citep{NIPS2016_afda3322,STREHL20081309}, \textit{optimism} 
\citep{NIPS2006_c1b70d96, NIPS2008_e4a6222c}, 
\textit{curiosity} 
\citep{NIPS2016_abd81528, pmlr-v70-pathak17a, pmlr-v100-schultheis20a}
, and \textit{reward shaping} \citep{Dann_Zambetta_Thangarajah_2019}.
Secondly, there are multi-environment approaches that incrementally learn a task across environments, for example, through \textit{transfer learning} \citep{weiss2016survey, parisi2021, 2020sequential, 2018abel, 2021lipschitz}, \textit{continual learning} \citep{Kirkpatrick3521}, \textit{meta-learning} \citep{finn2017model}, and \textit{curriculum learning} \citep{narvekar2020curriculum, luo2020accelerating}. We discuss these approaches below and contrast them to our contributions. 

\subsection{Single-environment}
The notion of intrinsic rewards for exploration originated with Schmidhuber \citep{schmidhuber1991}, who proposed encouraging exploration by visiting unpredictable states. 
In recent years, researchers in RL have extensively studied auxiliary rewards to compensate for the lack of external rewards.
Many intrinsic rewards have been proposed, including bonuses based on visitation counts and prediction errors \citep{NIPS2016_afda3322,STREHL20081309}.
For instance, a dynamic model can be learned 
to predict the next state \citep{NIPS2016_abd81528, pmlr-v70-pathak17a}.
Here, the agent is incentivized to explore unpredictable states by granting a bonus proportional to its prediction error.
Another idea, by Schultheis et al. \citep{pmlr-v100-schultheis20a}, is to maximize extrinsic rewards by meta-gradient to learn intrinsic rewards. Also, potential-based reward shaping to encourage agents to explore more has been studied as a way of increasing the learning speed \citep{Dann_Zambetta_Thangarajah_2019}.

These methods of exploration are often called agent-centric since they are based on updating the agent's belief, e.g., through the forward model error. In this sense, these works fall into the same category as ours; both rely on the hypothesis that exploration at the same time with exploitation by agents can be sufficient to achieve a (sub-) optimal policy. 
Another common assumption is that in all the papers following this method, the main goal and, consequently, the target award, i.e., the extrinsic reward is clear, and we have knowledge about it. 
The main difference between our work and others is that their purpose in the first place is only to explore as much as possible, and they are based on the assumption that by exploring more, the agent can learn an optimal policy. 
In contrast, we do not have such an assumption, and we do not seek to visit the states as much as possible. Instead, we focus only on intelligent and auxiliary exploration, which is resulted from transferred knowledge gained on intermediate partially solved tasks, which is in line with the main goal and accelerates the learning of the main task, not a complete acquaintance of the environment.

\subsection{Multi-environment}
%
The second set of papers reviewed followed the idea of incrementally learning tasks that have long been known in machine learning \citep{ring1994}.
In RL, although recent methods have mostly focused on policy and feature transfer, some have studied exploration transfer \citep{parisi2021}.
In policy transfer, behaviors are transferred to a new agent (student) by a pre-trained agent (teacher). 
The student is trained to minimize the KullbackLeibler divergence to the teacher using policy distillation, for example, \citep{DBLP}.
Other methods re-use policies from source tasks directly to create a student policy \citep{10.1145/1160633.1160762, NIPS2017_350db081}. On the other hand, a pre-learned state representation used in feature transfer encourages the agent to explore when it is presented with tasks \citep{Hansen2020Fast}.
In exploration transfer, in a task-agnostic and environment-centric scenario, the agent is first attempted by defining an intrinsic reward to visit the states as more as possible especially interesting states in multiple environments, instead of learning the main task, and then this prior knowledge is transferred as a bias to the main environment in which the main task is to be trained \citep{parisi2021}.
Studies on continual RL seek to determine how learning one or more tasks can accelerate learning of other tasks, as well as avoid ruinous forgetting \citep{Kirkpatrick3521}.
Meta RL and curriculum learning, on the other hand, focus on exploiting underlying similarities between tasks to speed up the learning process of new tasks \citep{finn2017model,narvekar2020curriculum, luo2020accelerating}. 
Many studies have been done under the Lifelong RL framework, where the tasks change following a specific distribution \citep{2020sequential, 2018abel, 2021lipschitz}. In \citep{2018abel, 2021lipschitz}, by spending a lot of computational costs, knowledge transfer is done through transferring initializers. Most are only compatible with ($\epsilon$, $\delta$)-PAC algorithms \citep{2020sequential, 2018abel, 2021lipschitz, strehl2009}

In the mentioned incrementally learning task methods, it is assumed that there are several tasks/environments that, with the help of the experience gained on one or multiple tasks/environments, we seek to accelerate the learning of the same task in the new environment or the new task in the same environment. 
As a result, the most important hypothesis is that tasks/environments are very similar to each other, which is not always consistent with reality and deprives us of the possibility of using these methods on any task and any environment. On the contrary, our proposed framework does not need to have several similar environments/tasks. In particular, by defining only one assistant reward along with a decreasing function, we successfully represent a large number of MDPs, which essentially does not require the assumption of switching tasks over a particular distribution. Therefore, it can be used for any environment and any task.
Another point is that in transfer learning, the learning process consists of two phases pre-training and the end-training phase, which are isolated from each other. More specifically, it is assumed that the learning on the previous tasks has been done thoroughly and with good accuracy. In our proposed framework, however, we are smoothly and consistently shifting between the auxiliary goals and the main goal. Not only do we not assume any tasks are already learned, but we also do not need to learn them completely and accurately, i.e., we only have one iteration of training per auxiliary task. 
In the other studies, like \citep{2020sequential}, which identify the closest task for knowledge transfer, there is a lot of computational cost and limit on the learning transfer method and the algorithms that can be combined with their methods. But our approach does not add computational complexity. Also, it can transfer knowledge through both value and policy transfer and is compatible with any algorithm.
Note that we have an integrated learning process so that when considering the acceleration of convergence, we do not neglect to consider the time required to learn the auxiliary goals.
The last issue is related to \citep{parisi2021}, where the authors try to explore as much as possible in the pre-training stage. On the contrary, we do not suppose such a hypothesis in our work. As a result, we only focus on defining auxiliary goals for faster learning of the task.

\section{Conclusion and Future Work} \label{section6}
Sample inefficiency is a critical weakness of current developments in RL. RL algorithms must become more efficient if they are to achieve their full potential in many real-world applications. In this paper, we introduce a new framework \texttt{TA-Explore} for smart exploration. The main idea is to use simpler assistant tasks that are aligned with the main task that gradually progress toward the main difficult task. Thus helping the agent in finding a more efficient learning trajectory.  
The \texttt{TA-Explore} framework is extremely flexible; it can be applied to any RL task, any type of environment. In addition, any type of RL algorithm can be used as its backbone. 
We conducted comprehensive experiments on a wide range of tasks with different difficulties, with different RL algorithms -both value-based methods and policy-based methods and both tabular methods and deep RL methods.
The results show the excellent performance of the proposed method in increasing the convergence speed.
Moreover, \texttt{TA-Explore} has no additional computational cost and complexity.
Therefore, our proposed framework, which owes its superiority to the definition of the teacher-assisted goal, can impact the field of RL and make researchers think about a better definition of the reward, relaxing limiting assumptions, and no need to complete learning of the prior tasks before the transfer learning process. 
Proving the sample complexity improvement in such a setting is potentially future work.
Also, for future work, adding a self-tuning feature to the $\beta$ function can go a long way in overcoming the only current limitation of the proposed method, which is to select the $\beta$ function experimentally. On the other hand, examining its performance in multi-agent environments under the partially observable Markov decision process (POMDP) assumption, which has higher complexity and suffers from slow learning of the objective function, can be considered a valuable task.

\section*{Acknowledgment}
This work was partially supported by the Swedish Research Council through grant agreement no. 2020-03607 and in part by 
Sweden's Innovation Agency (Vinnova). 
The computations were enabled by resources provided by the National Academic Infrastructure for Supercomputing in Sweden (NAISS) and the Swedish National Infrastructure for Computing (SNIC) 
at Chalmers Centre for Computational Science and Engineering (C3SE) partially funded by the Swedish Research Council through grant agreement no. 2022-06725 and no. 2018-05973.





\balance
\bibliographystyle{elsarticle-harv} 
\bibliography{sample_without_url.bib}

\end{document}